\documentclass{article}
\usepackage{amsmath,amsfonts,amsthm,algorithm,algorithmic,enumerate}

\usepackage[sort&compress]{natbib}

\usepackage{hyperref}

\bibpunct{(}{)}{;}{a}{,}{,}

\newcommand{\R}{\mathbb{R}}				    
\newcommand{\BP}{\begin{proof}}			    
\newcommand{\EP}{\end{proof}}			

\makeatletter
\floatstyle{\ALG@floatstyle}
\newfloat{protocol}{htbp}{loa}
\floatname{protocol}{Protocol}
\makeatother

\newtheorem{theorem}{Theorem}
\newtheorem*{theorem*}{Theorem}

\newtheorem{lemma}{Lemma}
\newtheorem{corollary}{Corollary}

\theoremstyle{definition}
\newtheorem{definition}{Definition}
\newtheorem*{remark}{Remark}

\begin{document}

\title{Aggregating Algorithm competing with Banach lattices}
\date{}

\author{Fedor Zhdanov, Alexey Chernov, and Yuri Kalnishkan\\
Computer Learning Research Centre,\\
Department of Computer Science,\\
Royal Holloway University of London,\\
Egham, Surrey, TW20 0EX, UK\\
\{fedor,chernov,yura\}@cs.rhul.ac.uk\\
}

\maketitle
\thispagestyle{empty}

\begin{abstract}
The paper deals with on-line regression settings with signals belonging to a Banach lattice. Our algorithms work in a semi-online setting where all the inputs are known in advance and outcomes are unknown and given step by step. We apply the Aggregating Algorithm to construct a prediction method whose cumulative loss over all the input vectors is comparable with the cumulative loss of any linear functional on the Banach lattice. As a by-product we get an algorithm that takes signals from an arbitrary domain. Its cumulative loss is comparable with the cumulative loss of any predictor function from Besov and Triebel-Lizorkin spaces. We describe several applications of our setting.
\end{abstract}

\section{Introduction}
In this paper we consider an online regression task.
A sequence of outcomes is predicted step by step.
In the beginning of each step
we are given a signal related to an outcome.
After we make our prediction,
the true outcome is announced.
We are interested to match a relationship between signals and their outcomes.
In a simple case each signal is an input vector of some variables
and this relationship is assumed to be linear;
linear regression minimising the expected loss is studied in statistics.
We assess the quality of predictions
by means of the loss accumulated over several trials.
This loss is compared against the loss of predictors from some benchmark class.
In this paper we prove the upper bounds for the cumulative losses of our algorithms in the form
\begin{equation}\label{eq:generalbound}
L_T \le L_T(\theta) + R(T,\theta)
\end{equation}
where the $T$ is the number of prediction step,
$L_T$ is the cumulative loss of an algorithm over $T$ steps,
$L_T(\theta)$ is the loss of any predictor $\theta$ from a chosen benchmark class over $T$ steps,
and $R(T,\theta)$ is an additional term called a regret term.
We say that our algorithm \emph{competes} with any function from a chosen benchmark class
if the order of $R(T,\theta)$ by $T$ is sublinear.

The case when the signal is an input vector of some variables is well studied in computer learning.
Many algorithms \citep[see][Chapter 10]{CesaBianchi2006}
compete with the benchmark class of linear functions of input vectors $\R^n \to \R$.
Some of them can be generalized to compete with
the benchmark class of all functions from a Reproducing Kernel Hilbert spaces \citep{Gammerman2004,VovkRKHS,Kivinen2004}.

The novelty of this paper
is in the expansion of the class of signals
to the signals from abstract normed vector spaces.
In this paper we consider Banach lattices,
they are Banach spaces with some additional structural assumptions.
The performance of our algorithm
is compared with the performance of any vector from a dual lattice
(so with a linear predictor on the signal).
We show that this framework can be useful when signals are digital images or sounds.

From one side,
by example of AAR \citep{VovkCOS} we show that algorithms developed to compete
with linear functions of a vector input can be slightly modified to work in our framework.
On the other side,
this surprising result comes with the assumption that all the input signals are known in advance.
We call it semi-online setting. We show that in some applications of online regression a semi-online algorithm does not appear as a drawback.

We modify our algorithm
to be able to work with finite-dimensional input vectors from a domain of $\R^m$
and the benchmark class of functions belonging to a Sobolev space.
This may give a wide spectrum of applications,
for example prediction of Brownian motion
(which almost surely can be said to belong to a Sobolev space, see \citealp{VovkCWPR}).

The paper is organized as follows.
In Section~\ref{sec:CWDF}
we give proofs of the theoretical bounds
for the performance of an algorithm taking finite-dimensional input vectors.
The benchmark class of predictors
is a class of linear functions of input vectors having non-euclidian norms.
Section~\ref{sec:TB} describes
the proof of the main theoretical bounds
for an algorithm working with Banach lattices.
In Section~\ref{sec:applications}
we describe several applications of our algorithms.
Section~\ref{sec:Discussion}
discusses some open problems
We include
some complicated proofs and algorithms in the Appendix.

\section{Competing with different norms}\label{sec:CWDF}
A game of prediction contains three components:
a space of outcomes $\Omega$,
a decision space $\Gamma$,
and a loss function $\lambda:\Omega\times\Gamma\to \mathbb{R}$.
We are interested in the square-loss game
with $\Omega = [-Y,Y], Y > 0$, $\Gamma = \mathbb{R}$,
and the loss function $\lambda(y,\gamma) = (y-\gamma)^2, y \in \Omega, \gamma \in \Gamma$.
The game of prediction is being played repeatedly by a learner
receiving some signals $x_t$ from a linear space $S$,
and follows the prediction protocol:

\begin{protocol}[H]
  \caption{Online regression}
  \label{prot:onlregr}
  \begin{algorithmic}
    \STATE $L_0:=0$.
    \FOR{$t=1,2,\dots$}
      \STATE Reality announces a signal $x_t \in S$.
      \STATE Learner announces $\gamma_t\in\Gamma$.
      \STATE Reality announces $y_t\in\Omega$.
      \STATE $L_t:=L_{t-1}+\lambda(y_t,\gamma_t)$.
    \ENDFOR
  \end{algorithmic}
\end{protocol}

Here $L_t$ is the cumulative loss of the learner.
We are interested in obtaining upper bounds
on the loss of the learner in the form~\eqref{eq:generalbound} for any $\theta \in S^*$.
The quality of the work of the learner
can be measured by the order of growth of the regret term in $T$.

We use the prediction method called Aggregating Algorithm for Regression (AAR)
developed initially \citep{VovkCOS} for the case $S = \R^n$.
It takes a parameter $a>0$
and gives its prediction of an outcome at a step $T$ by formula
\begin{equation*}
\gamma_T = \sum_{i=1}^{T-1} y_t x_t' \left(aI + \sum_{t=1}^T x_t x_t'\right)^{-1} x_T.
\end{equation*}
Here $I$ is $n \times n$ identity matrix.
It performs as well as any linear predictor $\theta$
(given an input $x$ (column vector),
a predictor $\theta$ predicts $\theta' x$).
It is known that
\begin{theorem*}[\citealp{VovkCOS}] 
  For all $a>0$,
  all positive integers $T$,
  all input vectors $x_1,x_2,\ldots x_T \in \R^n$ such that $||x_t||_\infty \le X, t=1,2,\ldots,T$,
  and all $\theta \in \R^n$,
  the loss of AAR satisfies
  \begin{equation}\label{eq:AAR}
  L_T(\text{AAR}) \le L_T(\theta) + a||\theta||^2_2 + nY^2 \ln
  \left(\frac{TX^2}{a} + 1\right).
  \end{equation}
\end{theorem*}

The regret term in this bound has a logarithmic order of growth in $T$
but it is linear in $n$. Therefore it is applicable for the case of
small dimension $n$ and large $T$. We shall now prove an upper bound
that grows slowly in $n$ and depends on non-euclidian norms in $\R^n$.
We use the constants of the norms equivalence.

\begin{lemma}\label{lem:eqcon}
  Let $a \in \mathbb{R}^n$, $1 \le p \le 2$, and $1/p+1/q=1$. Then
  \begin{gather*}
  \|a\|_2 \le \|a\|_p, \\
  \|a\|_2 \le n^{1/2-1/q}\|a\|_q.
  \end{gather*}
\end{lemma}
\BP
  The first inequality follows from the fact that the function $f(p) =
  \|a\|_p$ is decreasing in $p$. Indeed, for $p \ne 0$
  \begin{equation*}
  f'_p = \frac{-1}{p^2}\sum_{i=1}^n|a_i|^{p-1}  \left(\sum_{i=1}^n|a_i|^p
  \right)^{1/p-1}\le 0.
  \end{equation*}

  To prove the second inequality we consider the Holder
  inequality for $x,y \in \mathbb{R}^n$ and $b \ge 1$ \citep[see][p.21]{Beckenbach1961}:
  \begin{equation*}
  \sum_{i=1}^n |x_iy_i| \le \left( \sum_{i=1}^n |x_i|^b\right)^{1/b}
  \left( \sum_{i=1}^n |y_i|^c\right)^{1/c}
  \end{equation*}
  for $1/b+1/c=1$.
  This implies
  \begin{equation*}
  \|a\|_2^2 = \sum_{i=1}^n |a_i|^2 \le \left( \sum_{i=1}^n (|a_i|^2)^{q/2}\right)^{2/q} \left( \sum_{i=1}^n
  |1|^{\frac{q}{q-2}}\right)^{\frac{q-2}{q}},
  \end{equation*}
  for $b = q / 2 \ge 1$ and $c = \frac{q}{q-2}$. Therefore $\|a\|_2 \le
  n^{1/2-1/q}\|a\|_q$.
\EP

We denote the space of $n$-dimensional real vectors $x =
(x^1,\ldots,x^n)$ equipped with the $q$-norm $\|x\|_q =
\left(\sum_{i=1}^n(|x^i|)^{q}\right)^{1/q}$ by $\ell_q^n$, $q \ge 1$. Let $p$ be such that $1/p+1/q=1$.

\begin{lemma}\label{lem:AARp}
  For each positive integer $T$
  and any real positive $Y,X$
  there is a constant $a>0$
  such that for any sequence $(x_1,y_1),\ldots,(x_T,y_T)$ such that $||x_t||_q \le X, |y_t| < Y, t=1,2,\ldots,T$
  and all $\theta \in \ell_p^n$
  the loss of AAR with the parameter $a$ satisfies
  \begin{equation}\label{eq:AARp}
  L_T(\mathrm{AAR}) \le L_T(\theta) + (Y^2X^2 + \|\theta\|_p^2) T^{1/2}n^{1/2-1/\max(q,p)}\Theta.
  \end{equation}
\end{lemma}
\BP
  Following \citep[Theorem 3]{VovkRKHS} we get
  \begin{equation*}
  L_T(\mathrm{AAR}) \le L_T(\theta) + a\|\theta\|_2^2 + Y^2 T \frac{\max_{t=1,\ldots,T} \|x_t\|_2^2}{a}.
  \end{equation*}

  If $q \ge 2$,
  then by Lemma~\ref{lem:eqcon} $\|x_t\|_2^2 \le n^{1-2/q} \|x_t\|_q^2$
  and $\|\theta\|_2^2 \le \|\theta\|_p^2$.
  This leads to the regret term $a\|\theta\|_p^2 + \frac{Y^2 T n^{1-2/q} X^2}{a}$.
  By choosing $a = \sqrt{Tn^{1-2/q}}$
  we obtain the regret term $(Y^2X^2 + \|\theta\|_p^2) T^{1/2}n^{1/2-1/q}$.

  If $1 \le q \le 2$,
  then by Lemma~\ref{lem:eqcon} $\|x_t\|_2^2 \le \|x_t\|_q^2$
  and $\|\theta\|_2^2 \le n^{1-2/p}\|\theta\|_p^2$.
  This leads to the regret term $an^{1-2/p}\|\theta\|_p^2 + \frac{Y^2 T X^2}{a}$.
  For the same $a = \sqrt{Tn^{1-2/q}} = \sqrt{Tn^{2/p-1}}$
  we obtain the regret term $(Y^2X^2 + \|\theta\|_p^2) T^{1/2}n^{1/2-1/p}$.
\EP

\begin{remark} We can deduce another bound from~\eqref{eq:AAR}. We have $\|x\|_\infty \le \|x\|_q$ for any $q \ge 1$. Since $\|\theta\|_2 \le \|\theta\|_p$ if $p \le 2$
the upper bound becomes
\begin{equation*}
L_T(\mathrm{AAR}) \le L_T(\theta) + a\|\theta\|_p^2
+ Y^2 n \ln \left(\frac{TX^2}{a}+1\right)
\end{equation*}
for $q \ge 2$.
Since $\|\theta\|_2 \le n^{1/2-1/p}\|\theta\|_p$ if $p \ge 2$
the upper bound becomes
\begin{equation*}
L_T(\mathrm{AAR}) \le L_T(\theta) + an^{1/2-1/p}\|\theta\|_p^2
+ Y^2 n \ln \left(\frac{TX^2}{a}+1\right)
\end{equation*}
for $1 \le q < 2$.
The last bound is better in~$T$ but worse in~$n$ than~\eqref{eq:AARp}.
In our main theorem we consider spaces of infinite dimension. The role of~$n$ is played there by the dimension of the span of the inputs so far, which is generally~$T$, and only the bound similar to~\eqref{eq:AARp} remains nontrivial.
\end{remark}

Many researchers in machine learning consider kernel methods.
Some algorithms which use kernels
are able to compete with functions from a Reproducing Kernel Hilbert space.
Our abstract framework allows us
to formulate the upper bound on the loss of an algorithm
working in an abstract Hilbert space $S = H$.
We denote the scalar product in $H$ by $\langle \cdot,\cdot \rangle$.
The algorithm which we use is called KAAR (Kernelized AAR).
It takes a parameter $a>0$ and gives its prediction of an outcome at a step $T$ by formula
\begin{equation}\label{eq:KAARformula}
\gamma_T = (y_1,\ldots,y_{T-1},0) \left(aI + \widetilde{K}\right)^{-1} \widetilde{k}(x_T).
\end{equation}
Here $I$ is $T \times T$ identity matrix,
$\widetilde{K}$ is a matrix of mutual scalar products $\langle x_i, x_j \rangle, x_i \in H, i,j=1,\ldots,T$,
and $\widetilde{k}(x_T)$ is the last column of $\widetilde{K}$.
It performs as well as any linear predictor $h \in H$
(given an input $x$,
a predictor $h$ predicts $\langle h, x \rangle$).
The following theoretical bound for KAAR follows from Theorem 3 in \citet{VovkRKHS}.
\begin{theorem*}[KAAR theoretical bound]
For any $a>0$,
every positive integer $T$,
any sequence $(x_1,y_1),\ldots,(x_T,y_T), x_t \in H, |y_t| < Y, t=1,\ldots,T$,
and all $h \in H$,
the loss of KAAR satisfies
\begin{equation}\label{eq:KAARbound}
L_T(\text{KAAR}) \le L_T(h) + a||h||_H^2 + Y^2 \ln \det\left(I +
\frac{1}{a}\widetilde{K}\right).
\end{equation}
\end{theorem*}

\section{Theoretical bound for the algorithm competing with Banach lattices}\label{sec:TB}
In this section we need to consider a different protocol than Protocol~\ref{prot:onlregr}. The learner plays the game following semi-online Protocol~\ref{prot:theorem1}.
\begin{protocol}[h]
  \caption{Semi-online abstract regression}
  \label{prot:theorem1}
  \begin{algorithmic}
    \STATE $L_0:=0$.
    \STATE Reality announces number of steps $T$ and signals $x_1,\ldots,x_T \in S$.
    \FOR{$t=1,2,\dots,T$}
      \STATE Learner announces $\gamma_t\in \R$.
      \STATE Reality announces $y_t\in [-Y,Y]$.
      \STATE $L_t:=L_{t-1}+(y_t-\gamma_t)^2$.
    \ENDFOR
  \end{algorithmic}
\end{protocol}

He competes with all the functions from the dual space $S^*$. His algorithm BLAAR (Banach Lattices-competing AAR) working in $L_p(\mu), p \ge 1$ spaces is described as Algorithm~\ref{alg:lattices} and derived in the Appendix. Recall, that $L_p(\mu)$ is the space of all $\mu$-equivalent classes of $p$-integrable $\mu$-measurable functions on a $\mu$-measurable space $\mathbf{X}$:
$$\|f\|_{L_p(\mu)} = \left(\int_\mathbf{X} |f|^p d\mu\right)^{1/p}< \infty.$$
We use the notation $L_p = L_p(\mu)$.

\begin{algorithm}[h]
  \caption{BLAAR for $L_p$.}
  \label{alg:lattices}
  \begin{algorithmic}
    \STATE Reality announces number of steps $T$ and signals $x_1,\ldots,x_T \in L_p$.
    \STATE \textbf{Step 1.} Find the linearly independent subset of $x_1,\ldots,x_T$ with the maximum number of vectors: $x_{r_1},\ldots,x_{r_n}$.
    \STATE \textbf{Step 2.} Solve the following optimization problem. Maximize the absolute value of the determinant of a matrix $C = \{c_{ij}\}_{ij}$ of sizes $n \times n$: $|\det C| \to \max$ with a restriction
    \begin{equation*}
         \left\|\sqrt{\sum_{i=1}^n |\gamma_i|^2}\right\|_{L_p} \le 1, \text{ where }\gamma_i = \sum_{j=1}^n c_{ij} x_{r_j}.
    \end{equation*}

    Let the matrix $D$ be the inversion of the matrix $C$: $D= C^{-1}$.
    \STATE \textbf{Step 3.} Take $a=\sqrt{T n^{-|1/2-1/p|}}$. Use it as a parameter for KAAR.
    \FOR{$t=1,2,\dots,T$}
        \STATE Let $x_s = \sum_{i=1}^n \alpha_{si} x_{r_i}$ for $s = 1,\ldots,T$. Apply KAAR for prediction at each step by formula \eqref{eq:KAARformula}. In the matrix of scalar products use
        \begin{equation}\label{eq:scaltheta}
            \widetilde{K}_{sl} = \frac{1}{n}\sum_{i,j=1}^{n} \alpha_{si}\alpha_{lj} \sum_{k=1}^n d_{ik} d_{jk},\qquad s,l = 1,\ldots,T.
        \end{equation}
    \ENDFOR
  \end{algorithmic}
\end{algorithm}

We prove the following upper bound for the cumulative loss of BLAAR.
It performs as well as any linear predictor $f \in (L_p)^*$
(given a signal $x$,
a predictor $f$ predicts $f(x)$).
\begin{theorem}\label{thm:BLAARLp}
  Suppose we are given $p > 1$
  and $x_1,\ldots,x_T \in L_p$
  for any positive integer $T$.
  Assume also that $\|x_t\| \le X$
  and $|y_t| \le Y$
  for all $t=1,\ldots,T$.
  Then there exists $a > 0$ such that
  for all $f \in (L_p)^*$
  and any sequence $y_1,\ldots,y_T$ we have
  \begin{equation}\label{eq:BoundBLAARLp}
  L_T(\mathrm{BLAAR(a)}) \le L_T(f) + (Y^2X^2 + \|f\|^2)T^{1/2 + |1/2-1/p|}.
  \end{equation}
\end{theorem}
The proof of this theorem is given in the Appendix.
The main argument is based on Corollary 5 in~\citet{Lewis1978}.
Note that if in Lemma~\ref{lem:AARp} we take $n=T$,
then AAR gives the regret term of the same order $T^{1-1/p}$ for $\ell_p^n, p \ge 2$.

It is possible to generalize the result for Banach lattices of more general type.
The algorithm becomes rather tricky
because it is based on the complex interpolation method,
and we do not discuss it here.
We formulate the theorem for the theoretical bound of this algorithm.
First we give the definition of a Banach lattice \citep[see]{Lindenstrauss1979}.
\begin{definition}{
  \rm \emph{A Banach-lattice} is a partially ordered Banach space $B$ over the reals provided
  \begin{enumerate}[(i)]
    \item $x \le y$ implies $x+z \le y+z$, for every $x,y,z \in B$,
    \item $ax \le 0$ for every $x \le 0$ in $B$ and every nonnegative real $a$,
    \item for all $x,y \in B$ there exists a least upper bound $x \vee y$ and a greatest lower bound $x \wedge y$,
    \item $\|x\| \le \|y\|$ whenever $|x|\le|y|$, where the absolute value of $|x|$ of $x \in B$ is defined by $|x|=x \vee (-x)$.
  \end{enumerate}
}\end{definition}
The lattices are a well-studied wide class of Banach spaces.
For example, any $L_p(\mu)$ is a lattice
(consequently, $\ell_p$ is a lattice).
Other examples of Banach lattices are Orlicz spaces.
Another more intuitive definition of a Banach lattice is given in~\citet{Tomczak1989}:
\begin{definition}{
  \rm If $(\Omega,\Sigma,\mu)$ is a measure space then
  a Banach space $B$ is called \emph{a Banach-lattice on $(\Omega,\Sigma,\mu)$}
  if $B$ consists of equivalence classes of $\mu$-measurable real functions on $\Omega$ such that
  if $f$ is $\mu$-measurable,
  $g \in B$,
  and $|f|\le |g|$ $\mu$-a.e.,
  then $f \in B$ and $\|f\| \le \|g\|$.
}\end{definition}
We will use some pointwise expressions with elements of Banach lattices, e.g.,
$z = \left(\sum_j |f_j|^p \right)^{1/p}$, $1 \le p < \infty$,
where $\{f_j\}$ is a finite sequence in $B$.
Our main theorem uses the following structural properties of Banach lattices.
They are similar to convexity properties of standard Banach spaces.
\begin{definition}{
  \rm Assume $1 \le p, q \le \infty$ and $B$ is a Banach lattice.
  \begin{enumerate}[(i)]
    \item $B$ is called \emph{$p$-convex} if
        there exists a constant $M$ so that
        $\forall n, \forall x_1,\ldots,x_n \in B$
        \begin{equation*}
        \left\|\left(\sum_{i=1}^n |x_i|^p\right)^{1/p}\right\| \le M\left(\sum_{i=1}^n \|x_i\|^p\right)^{1/p}.
        \end{equation*}
        The smallest possible value of $M$ is denoted by $M^{(p)}(B)$.
    \item $B$ is called \emph{$q$-concave} if
        there exists a constant $M$ so that
        $\forall n, \forall x_1,\ldots,x_n \in B$
        \begin{equation*}
        \left(\sum_{i=1}^n \|x_i\|^q\right)^{1/q} \le M\left\|\left(\sum_{i=1}^n |x_i|^q\right)^{1/q}\right\|.
        \end{equation*}
        (with the usual convention for $q=\infty$). The smallest possible value of $M$ is denoted by $M_{(q)}(B)$.
  \end{enumerate}
}\end{definition}
Every Banach lattice is $1$-convex and $\infty$-concave.
As a non-trivial example,
the space $L_p(\mu)$ is a $p$-convex and $p$-concave Banach lattice
with $M^{(p)}(L_p)=M_{(q)}(L_p)=1$
(this can be easily verified).
If $p \ge 2$,
we can think it is $p$-concave and $2$-convex,
and if $1 \le p < 2$
we can think it is $p$-convex and $2$-concave
\citep[see][Proposition 1.d.5]{Lindenstrauss1979}.

\begin{theorem}\label{thm:BLAAR}
  Let $B$ be a $p$-convex and $q$-concave Banach lattice $B$,
  $1 < p \le 2 \le q < \infty$.
  Suppose we are given $x_1,\ldots,x_T \in B$ for any positive integer $T$.
  Assume also that $\|x_t\| \le X$ and $|y_t| \le Y$ for all $t=1,\ldots,T$.
  Then there exists an algorithm taking some $a > 0$ such that
  for all $f \in B^*$ and any sequence $y_1,\ldots,y_T$ we have
  \begin{equation}\label{eq:BoundBLAAR}
  L_T(\mathrm{Algorithm(a)}) \le L_T(f) + (Y^2X^2 + \|f\|^2)M^{(p)}(B)M_{(q)}(B)T^{1/2+\alpha},
  \end{equation}
  where $\alpha = \max\{\frac{1}{p}-\frac{1}{2},\frac{1}{2}-\frac{1}{q}\}$, and $s > 0$.
\end{theorem}
The proof of this theorem is similar to the proof of Theorem~\ref{thm:BLAARLp}.
The main argument bases on Theorem 28.6 in~\citet{Tomczak1989} or Corollary 3.5 in~\citet{Pisier1979},
though their proof techniques are different
from the proof technique of the main argument for Theorem~\ref{thm:BLAARLp}.
The sequence of steps in the algorithm
follows the steps of the proof of these theorems.

\section{Applications}\label{sec:applications}
In this section we consider different applications of our main theorem.
They use Theorem~\ref{thm:BLAARLp} rather than Theorem~\ref{thm:BLAAR},
so the algorithm used to give predictions is Algorithm~\ref{alg:lattices}.

\subsection{Algorithm competing with functional Banach spaces}\label{ssec:ACBS}
A different protocol than Protocol~\ref{prot:theorem1}
is usually considered in the online regression literature:
inputs are elements of some domain $\mathbf{X} \subseteq \mathbb{R}^m$.
The goal is to find an algorithm
competing with all the functions from
a functional Banach space $\mathcal{B}$ on this domain $\mathbf{X}$.
Many algorithms are capable to compete with Reproducing Kernel Hilbert spaces.
The generalization of the notion of these spaces for the Banach case
is called a Proper Banach Functional space~\citep{VovkCWPR}.
\begin{definition}
  A \em{Proper Banach Functional space} (PBFS) on a set $\mathbf{X}$
  is a Banach space $\mathcal{B}$ of real-valued functions on $\mathbf{X}$ such that
  the evaluation functional $\varphi: f \in \mathcal{B} \mapsto f(x)$ is continuous
  for each $x \in \mathbf{X}$.
  We will use the notation $c_\mathcal{B}(x)$ for the norm of this functional:
  $c_{\mathcal{B}}(x) := \sup_{f:\|f\|_{\mathcal{B}} \le 1} |f(x)|$
  and for the embedding constant
  \begin{equation*}
  c_{\mathcal{B}} := \sup_{x \in \mathbf{X}} c_{\mathcal{B}}(x)
  \end{equation*}
  assumed to be finite.
\end{definition}
We will show further examples of PBFS with finite constant $c_{\mathcal{B}}$.
We state here that it is possible to apply BLAAR
to get the following upper bound in the standard protocol.
It performs as well as any predictor $f$ from a Banach lattice
which has the PBFS property
(given an input vector $x \in \R^m$,
a predictor $f$ predicts $f(x)$).
\begin{theorem}\label{thm:BLAAR2}
  Let $\mathbf{X}$ be an arbitrary set
  and $\mathcal{B}$ be a PBFS on $\mathbf{X}$
  and a $q$-convex and $p$-concave Banach lattice,
  $1 < q \le 2 \le p < \infty$.
  Suppose we are given $x_1,\ldots,x_T \in \mathbf{X}$ for any positive integer $T$.
  Assume also that $|y_t| \le Y$ for all $t=1,\ldots,T$.
  Then it is possible to apply BLAAR with a parameter $a > 0$
  such that for all $f \in \mathcal{B}$ and any sequence $y_1,\ldots,y_T$ we have
  \begin{equation}\label{eq:BoundBLAAR2}
  L_T(\mathrm{BLAAR}) \le L_T(f) + (Y^2c_{\mathcal{B}}^2+\|f\|^2)M_{(p)}(\mathcal{B})M^{(q)}(\mathcal{B})T^{1/2+\beta},
  \end{equation}
  where $\beta = \max\{\frac{1}{q}-\frac{1}{2},\frac{1}{2}-\frac{1}{p}\}$.
\end{theorem}
The proof of this theorem bases on the correspondence between $\mathbf{X}$ and $(\mathcal{B})^{**}$ and Theorem~\ref{thm:BLAAR} and almost fully repeats the proof of Corollary~\ref{cor:Sobolev} (follows further).

The regret term in~\eqref{eq:BoundBLAAR2} reaches its minimum by $p,q$ when $p=q=2$. In this case $\mathcal{B}$ is a Hilbert space. The PBFS property implies that $\mathcal{B}$ is a Reproducing Kernel Hilbert Space. In this case, the regret term is of order $T^{1/2}$ and coincides with the order of the regret terms given by the algorithms previously applied for competing with RKHS.

We can not apply Theorem~\ref{thm:BLAAR2} to $L_p$ spaces since they are not proper. But this theorem covers very important classes of Banach spaces: Besov and Triebel-Lizorkin spaces with appropriate parameters~\citep{Triebel1978}. We start our description with the discussion of the algorithm competing with fractional Sobolev spaces.

The main trick used in order to compete with Sobolev spaces is to identify each element of them with some element from $L_p$ of the same (up to a constant) norm and thus to impose a lattice structure on these spaces. This isomorphism can be found if $\mathbf{X}$ is an open, non-empty subset of $\mathbb{R}^m$ such that there exists a linear extension operator (see definition on p.1372 of \citealp{Pelczynski2003a}) from a Sobolev space $W_p^{s}(\mathbf{X})$ into $W_p^{s}(\mathbb{R}^m)$. This condition holds for Lipschitz domains 
(see, e.g., \citealp{Rogers2006}). We will further assume our domain is a Lipschitz domain.

Let us take a function $u(x): \mathbb{R}^m \to \mathbb{R}$ and by
\begin{equation*}
\widehat{u}(y) = \frac{1}{(2\pi)^{m/2}}\int_{\mathbb{R}^m} u(x)e^{-ixy} dx
\end{equation*}
denote a Fourier transform of $u(x)$. By $f^\vee$ we denote an inverse Fourier transform
\begin{equation*}
f^\vee(x) = \frac{1}{(2\pi)^{m/2}}\int_{\mathbb{R}^m} f(y)e^{ixy} dy
\end{equation*}
for a function $f$. Then the isomorphism between a Sobolev space and a subspace of $L_p$ is described by the following theorem. It is constructed using Bessel potentials.
\begin{theorem*}[Isomorphism of $W_p^{s}$ and $L_p$]
  Let $1 < p < \infty, s > 0$ such that $sp > m$. Then $W_p^s$ may be described as
  \begin{equation}\label{eq:convol}
  W_p^s = \{ f \in S'(\mathbb{R}^m): \left((1+\|y\|_2^2)^{s/2}\widehat{f}(y)\right)^\vee \in L_p(\mathbb{R}^m)\},
  \end{equation}
  where $S'(\mathbb{R}^m)$ is a collection of all tempered distributions on $\mathbb{R}^m$.
\end{theorem*}
The mapping in \eqref{eq:convol} means the convolution of a given function $f$ and the function with a polynomial Fourier transform $(1+\|y\|_2^2)^{s/2}$ (the latter called Bessel potential). Explicit expressions of these functions see in~\citet{Aronszajn1963}.

For the Sobolev spaces with $p=2$ and $s$ is an integer number, the proof is easy and based on the Plancherel's theorem of norm equivalence. For all $s,p$ see \cite{Triebel1992}, Theorem 1.3.2.

We can apply this theorem to get an algorithm competing with Sobolev spaces. This algorithm can be derived from the proof of the following corollary. 
\begin{corollary}\label{cor:Sobolev}
  Assume $\mathbf{X}$ is a Lipschitz domain, and $W_p^s(\mathbf{X})$ is a fractional Sobolev space of functions on $\mathbf{X}$, $s > 0, p > 1$. Suppose we are given $x_1,\ldots,x_T \in \mathbf{X}$ for any positive integer $T$. Assume also that $|y_t| \le Y$ for all $t=1,\ldots,T$. Then there exists an algorithm taking some $a > 0$ such that for all $f \in W_p^s(\mathbf{X})$ and any sequence $y_1,\ldots,y_T$ we have
  \begin{equation*}
  L_T(\mathrm{BLAAR}) \le L_T(f) + (Y^2c_{W_p^s}^2 + \|f\|^2) K T^{1/2 + |1/2-1/p|}.
  \end{equation*}
  Here $K$ is defined by isomorphism between $W_p^s$ and $L_p$.
\end{corollary}
The proof of this corollary is given in the Appendix.

Lately Besov $B_{p,q}^s$ and Triebel-Lizorkin $F_{p,q}^s$ function spaces begin to interest researchers due to their connections with wavelets theory. They have the PBFS property \citep[see][Proposition 7(ii)]{Triebel2005}, and $c_{B_{p,q}^s}, c_{F_{p,q}^s} < \infty$. By the embedding theorem \citep[Theorem 2.3]{Triebel1978} $F_{p,q}^s \to B_{p,q}^s \to F_{p,2}^{s'} = W_{p}^{s'}, 1 < p \le q < \infty, s > s'$. Here embedding $A \to B$ means there exists a constant $C$ and linear operator $T: A \to B$ such that for any $f \in A$ we have $Tf \in B$ and $\|Tf\|_B \le C \|f\|_A$. For Slobodetsky spaces $B_p^s = B_{p,p}^s$ we can use another result from the same theorem: $B_p^s \to W_p^s, 2 \le p < \infty, s > 0$. It helps to keep the parameter $s$ and thus do not increase constants in the regret term.
Using Corollary~\ref{cor:Sobolev} we can get the theoretical bound
\begin{equation*}
L_T(\mathrm{BLAAR}) \le L_T(f) + (Y^2c_{\{B,F\}_{p,q}^s}^2 + \|f\|_{\{B,F\}_{p,q}^s}^2) CT^{1/2+|1/2-1/p|}
\end{equation*}
for all $f \in \{B,F\}_{p,q}^s, p>1$ and some $C > 0$ defined as a multiplication of the embedding constants. An important benchmark class is the class of H\"{o}lder-Zygmund spaces $\mathcal{C}^s = B^s_\infty$ which is embedded to $B^{s'}_p = B^{s'}_{p,p}$ whenever $s' < s$. It is known that fractional Brownian motion $B^{(h)}$ almost surely belongs to $\mathcal{C}^s$, $s < h$.

\subsection{Application of the abstract framework}\label{ssec:AbstactFramewAppl}
In this section we describe an example
how our algorithm can be used in signal processing.
A signal can often be interpreted as a function on some domain, e.g.,
a picture can be thought of as
a mapping from points to colors.
A musical fragment can be thought of as
a mapping from a point in time into sound frequencies.
We may be given weak regularity restrictions
on the class these functions form, e.g.,
it can be a Sobolev or Besov space.
The family of Hilbert spaces is reasonably wide,
but if lacks many classes of functions of irregular behavior.

Imagine we are given a film consisting frames of resolution $1024 \times 768$ and
we want to predict some score calculated from each image.
The correct linear score for each image is given to us
only after we make a prediction about the score of this image.
Applying the algorithm from Lemma~\ref{lem:AARp} for prediction
we can get the following upper bound for the square loss of our predictions
\begin{equation*}
L_T(\mathrm{AAR}) \le (Y^2X^2 + \|\theta\|_p^2) T^{1/2}n^{1/2-1/p}
\end{equation*}
where $p \ge 2$,
$n = 1024 \times 768 = 786432$,
$T$ is the length of the film in frames,
$X$ is the maximal $q$-norm of images ($1/q+1/p=1$), and
$Y$ is the upper bound on the absolute value of the score.
The upper bound from the remark is worse in $n$,
and in our example $n$ is the dominating constant for reasonable films length.

On the other hand, the algorithm from Theorem~\ref{thm:BLAARLp} has the upper bound
\begin{equation*}
L_T(\mathrm{BLAAR}) \le (Y^2X^2 + \|\theta\|_p^2) T^{1-1/p}.
\end{equation*}
Then if we want to predict 24 frames per second (say, to detect defective frames),
the upper bound on the loss of the second algorithm
will be better if we work with films of duration less than $32768$ (around $9$ hours).
The higher the resolution of the images is the more advantage the second algorithm has.
This improvement is due to the fact that
it finds linearly independent vectors and significantly depends only on them.
Note that the example above works well in the semi-online setting.

\subsection{Learning a classifier}\label{ssec:learnclass}
Online regression algorithms are often applied in the batch setting, when one has a training set with input vectors and their labels and a test set containing just input vectors. In this case the semi-online setting does not appear as a drawback.

Online regression methods can be used to learn a linear classifier, for example Perceptron. \citet{CesaBianchi2005} use the AAR algorithm steps to make an algorithm to train a Perceptron and to derive upper bounds on the number of mistakes. They consider both linear classification and classification in an RKHS. We show that the combination of our preprocessing steps and their algorithm allows us to learn a classifier working in a PBFS. The abstract protocol may be considered here, but we describe the standard protocol to give the reader the better understanding of how our algorithms can be applied to classify vectors.
Let $(x_1,y_1),\ldots,(x_T,y_T)$ be a set of examples,
where $x_i \in \R^m$ is an input vector and
$y_i = \{-1,1\}$ is its label, $i=1,\ldots,T$.
The label corresponds to the class of the input vector.
Define the hinge loss $D_\gamma(f,(x,y)) = \max\{\gamma - y f(x)\}, \gamma > 0$
of any function $f$ from a Sobolev space $W_p^s, s >0, p > 1$.
If we make preprocessing steps described in the proof of Corollary~\ref{cor:Sobolev}
we will get vectors $r_1,\ldots,r_T \in \ell_2^n$
corresponding to our input vectors, for some $n \le T$.
At step $t$ the second-order Perceptron Algorithm \citep[see][Figure 3.1]{CesaBianchi2005} predicts
\begin{equation*}
\widehat{y}_t = \text{sign}\left[\left(\sum_{i \in \mathcal{M}_t} y_i r_i \right)'(aI_n + \sum_{i \in \mathcal{M}_t} r_i r_i')^{-1} r_t\right],
\end{equation*}
where $\mathcal{M}_t \subseteq \{1,2,\ldots\}$ is
the set of indices of mistaken trials ($y_i \ne \widehat{y}_i, i \in \mathcal{M}_t$)
before the step $t$.
It is possible to prove the following upper bound on the number of mistakes
\begin{theorem}
  It is possible to run the second-order Perceptron Algorithm
  on any finite sequence $(x_1,y_1),\ldots,(x_T,y_T)$ of examples such that
  the number $k$ of mistakes satisfies
  \begin{equation*}
  k \le \inf_{\gamma > 0} \min_{f \in W_p^s}\left(\frac{R(f,T,a)^2}{2\gamma^2}
  + \frac{D_\gamma(f)}{\gamma}
  + \frac{R(f,T,a)}{\gamma}\sqrt{\frac{D_\gamma(f)}{\gamma}
  + \frac{R(f,T,a)^2}{4\gamma^2}}\right)
  \end{equation*}
  where $R^2(f,T,a) = c_{W_p^s}^2(T^{1/2-1/p}\|f\|_{W_p^s}^2 + \frac{1}{a}\sum_{i \in \mathcal{M}_T} f(x_i)^2)$
  and $D_\gamma(f)$ is the cumulative hinge loss $\sum_{i=1}^T D_\gamma(f,(x_i,y_i))$ of $f$.
\end{theorem}
Note that $T^{1/2-1/p}$ in the theorem above is equal to the maximum number of linearly independent inputs (converted to the dual space $(W_p^s)^{**}$), so if the algorithm is run on the same sequence of inputs several times then this number remains the same.

\section{Discussion}\label{sec:Discussion}
The idea of competing with Banach spaces is not new. Vladimir Vovk \citep{VovkCWPR,VovkME} considers two different ways to do this. The first technique is based on the game-theoretic probability theory \citep{Shafer2001} and called Defensive Forecasting.
The second technique is based on
the metric entropy of the space
with which the learner wishes to compete.
The Aggregating Algorithm is used for prediction.
Suppose that input vectors are taken from a domain $\mathbf{X}\subseteq\mathbb{R}^m$.
The main difference in the theoretical bounds for two algorithms
can be described by an example of Slobodetsky spaces $B^s_p(\mathbf{X}) = B^s_{p,p}(\mathbf{X})$.
We always assume that $sp > m$:
this condition ensures that the elements of $B^s_p$ are continuous functions on $\mathbf{X}$
(see, e.g., \citealp{Triebel1978}).
Assuming $p\ge2$,
the known upper bound on the regret term is of order $O(T^{1-1/p})$
when the learner uses either the Defensive Forecasting
or our algorithm from Corollary~\ref{cor:Sobolev} to predict the outcomes.
This order does not depend on $s$.
The order $O(T^{m /(m+s)})$ is provided by Metric Entropy technique.
This order does not depend on $p$ and so this algorithm can be applied to compete with spaces with $p=1$.

The question asked by \citet{VovkME} is whether it is possible to create an algorithm which will involve both $p$ and $s$ parameters in the order of $T$ in the regret term.
Our paper gives another way to apply the Aggregating Algorithm and the order of the regret term corresponds to the order given by Defensive Forecasting: $T^{1-1/p}$. Thus we reduced the problem to the analysis of two different ways of using the Aggregating Algorithm to mix functions from Banach spaces. 

Our paper shows the same order of the regret term by $T$ as the Defensive Forecasting method. It allows us to think that this order may be optimal by $p$. The lack of lower bounds for methods which are capable to compete with functions from Banach spaces does not allow us to make a strong argument.

\section{Acknowledgements}\label{sec:acknowledgement}
We are grateful for useful comments to Vladimir Vovk.
This work has been supported by EPSRC grant EP/F002998/1
and ASPIDA grant
from the Cyprus Research Promotion Foundation.

\bibliographystyle{plainnat}

\begin{thebibliography}{21}
\providecommand{\natexlab}[1]{#1}
\providecommand{\url}[1]{\texttt{#1}}
\expandafter\ifx\csname urlstyle\endcsname\relax
  \providecommand{\doi}[1]{doi: #1}\else
  \providecommand{\doi}{doi: \begingroup \urlstyle{rm}\Url}\fi

\bibitem[Aronszajn et~al.(1963)Aronszajn, Mulla, and Szeptycki]{Aronszajn1963}
N.~Aronszajn, F.~Mulla, and P.~Szeptycki.
\newblock On spaces of potentials connected with {$L\sp{p}$} classes.
\newblock \emph{Ann. Inst. Fourier (Grenoble)}, 13:\penalty0 211--306, 1963.
\newblock ISSN 0373-0956.

\bibitem[Beckenbach and Bellman(1961)]{Beckenbach1961}
Edwin~F. Beckenbach and Richard Bellman.
\newblock \emph{Inequalities}.
\newblock Springer, Berlin, 1961.

\bibitem[Cesa-Bianchi et~al.(2005)Cesa-Bianchi, Conconi, and
  Gentile]{CesaBianchi2005}
Nicol{\`o} Cesa-Bianchi, Alex Conconi, and Claudio Gentile.
\newblock A second-order perceptron algorithm.
\newblock \emph{SIAM Journal on Computing}, 34:\penalty0 640--668, 2005.

\bibitem[Cesa-Bianchi and Lugosi(2006)]{CesaBianchi2006}
Nicol{\`o} Cesa-Bianchi and G{\'a}bor Lugosi.
\newblock \emph{Prediction, learning, and games}.
\newblock Cambridge University Press, Cambridge, UK, 2006.

\bibitem[Gammerman et~al.(2004)Gammerman, Kalnishkan, and Vovk]{Gammerman2004}
Alexander Gammerman, Yuri Kalnishkan, and Vladimir Vovk.
\newblock On-line prediction with kernels and the complexity approximation
  principle.
\newblock In \emph{UAI}, pages 170--176, 2004.

\bibitem[John(1948)]{John1948}
Fritz John.
\newblock Extremum problems with inequalities as subsidiary conditions.
\newblock In \emph{Courant Anniversary Volume}, pages 187--204. Interscience
  Publishers, Inc., New York, N. Y., 1948.

\bibitem[Kivinen and Warmuth(2004)]{Kivinen2004}
Jyrki Kivinen and Manfred~K. Warmuth.
\newblock Online learning with kernels.
\newblock \emph{IEEE Transactions on Signal Processing}, 52:\penalty0
  2165--2176, 2004.
\newblock ISSN 1053-587X.

\bibitem[Lewis(1978)]{Lewis1978}
D.~R. Lewis.
\newblock Finite dimensional subspaces of {$L\sb{p}$}.
\newblock \emph{Studia Math.}, 63:\penalty0 207--212, 1978.
\newblock ISSN 0039-3223.

\bibitem[Lindenstrauss and Tzafriri(1979)]{Lindenstrauss1979}
Joram Lindenstrauss and Lior Tzafriri.
\newblock \emph{Classical {B}anach spaces. {II}}, volume~97 of \emph{Ergebnisse
  der Mathematik und ihrer Grenzgebiete [Results in Mathematics and Related
  Areas]}.
\newblock Springer, Berlin, 1979.

\bibitem[Pe{\l}czy{\'n}ski and Wojciechowski(2003)]{Pelczynski2003a}
Aleksander Pe{\l}czy{\'n}ski and Micha{\l} Wojciechowski.
\newblock Sobolev spaces.
\newblock In \emph{Handbook of the geometry of {B}anach spaces, {V}ol.\ 2},
  pages 1361--1423. North-Holland, Amsterdam, 2003.

\bibitem[Pisier(1979)]{Pisier1979}
G.~Pisier.
\newblock Some applications of the complex interpolation method to {B}anach
  lattices.
\newblock \emph{J. Analyse Math.}, 35:\penalty0 264--281, 1979.

\bibitem[Rogers(2006)]{Rogers2006}
Luke~G. Rogers.
\newblock Degree-independent {S}obolev extension on locally uniform domains.
\newblock \emph{J. Funct. Anal.}, 235:\penalty0 619--665, 2006.

\bibitem[Shafer and Vovk(2001)]{Shafer2001}
Glenn Shafer and Vladimir Vovk.
\newblock \emph{Probability and finance: It's only a game!}
\newblock Wiley Series in Probability and Statistics. Financial Engineering
  Section. Wiley-Interscience, New York, 2001.

\bibitem[Tomczak-Jaegermann(1989)]{Tomczak1989}
Nicole Tomczak-Jaegermann.
\newblock \emph{Banach-{M}azur distances and finite-dimensional operator
  ideals}, volume~38 of \emph{Pitman Monographs and Surveys in Pure and Applied
  Mathematics}.
\newblock Longman Scientific \& Technical, Harlow, 1989.

\bibitem[Triebel(2005)]{Triebel2005}
H.~Triebel.
\newblock Sampling numbers and embedding constants.
\newblock \emph{Tr. Mat. Inst. Steklova}, 248:\penalty0 275--284, 2005.

\bibitem[Triebel(1978)]{Triebel1978}
Hans Triebel.
\newblock \emph{Interpolation theory, function spaces, differential operators},
  volume~18 of \emph{North-Holland mathematical library}.
\newblock Amsterdam: North-Holland Pub. Co., 1978.

\bibitem[Triebel(1992)]{Triebel1992}
Hans Triebel.
\newblock \emph{Theory of function spaces. {II}}, volume~84 of \emph{Monographs
  in Mathematics}.
\newblock Birkh\"auser Verlag, Basel, 1992.

\bibitem[Vovk(2001)]{VovkCOS}
Vladimir Vovk.
\newblock Competitive on-line statistics.
\newblock \emph{International Statistical Review}, 69:\penalty0 213--248, 2001.

\bibitem[Vovk(2006{\natexlab{a}})]{VovkME}
Vladimir Vovk.
\newblock Metric entropy in competitive on-line prediction.
\newblock \emph{Technical report, arXiv:cs/0609045v1 [cs.LG], arXiv.org e-Print
  archive}, 2006{\natexlab{a}}.

\bibitem[Vovk(2006{\natexlab{b}})]{VovkRKHS}
Vladimir Vovk.
\newblock On-line regression competitive with reproducing kernel {H}ilbert
  spaces.
\newblock In \emph{Theory and applications of models of computation}, volume
  3959 of \emph{L.N.C.S.}, pages 452--463. Springer, Berlin,
  2006{\natexlab{b}}.

\bibitem[Vovk(2007)]{VovkCWPR}
Vladimir Vovk.
\newblock Competing with wild prediction rules.
\newblock \emph{Machine Learning}, 69:\penalty0 193--212, 2007.

\end{thebibliography}

\section*{Appendix}\label{sec:Appendix}
\subsection*{Proof of Theorem~\ref{thm:BLAARLp}}
The proofs of Theorems~\ref{thm:BLAARLp} and~\ref{thm:BLAAR}
base on the possibility to construct
an isomorphism between a finite-dimensional Banach lattice and a Hilbert space
such that the norms of vectors do not increase too much.
Precisely \citep[see][Theorem 28.6]{Tomczak1989},
\begin{theorem*}[Distance]\label{thm:BMDist}
  Let $B$ be a $p$-convex and $q$-concave Banach lattice,
  and $\mathcal{X}$ be an $n$-dimensional subspace of $B$.
  If $1 < p \le 2 \le q < \infty$
  then there exists an isomorphic operator $U: \mathcal{X} \to \ell_2^n$ such that
  \begin{equation}\label{eq:BMDist}
  \inf_{U} \|U\|\|U^{-1}\| \le n^\alpha M^{(p)}(B)M_{(q)}(B).
  \end{equation}
  Here $\alpha = \max\{\frac{1}{p}-\frac{1}{2},\frac{1}{2}-\frac{1}{q}\}$,
  $\|U\| = \sup_{x \in \mathcal{X}} \frac{\|Ux\|}{\|x\|}$,
  $\|U^{-1}\| = \sup_{r \in \ell_2^n} \frac{\|U^{-1}r\|}{\|r\|}$
  and infimum is taken over all isomorphisms between $\mathcal{X}$ and $\ell_2^n$.
\end{theorem*}
This theorem is formulated for $L_p, 1 < p < \infty$ spaces
as Corollary 5 in~\citet{Lewis1978}.
The infimum in this case is bounded by $n^{|1/2-1/p|}$.
The expression $\inf_{U} \|U\|\|U^{-1}\|$ defines
a \emph{Banach-Mazur} distance $d(\mathcal{X},\ell_2^n)$
between $\mathcal{X}$ and $\ell_2^n$.
For the case of a general Banach space
the John theorem~\citep{John1948} states that
$d(\mathcal{X},\ell_2^n) \le \sqrt{n}$ for any $n$-dimensional subspace $\mathcal{X} \in U$,
where $U$ is some Banach space.
Thus our theorems can be applied for the cases $p=1$ and $p=\infty$
though the regret term becomes trivial (of order $T$).
\begin{remark}
Interestingly, if one wants to omit the lattice structure on a Banach space, it is possible to prove a weaker result than the Distance theorem. For the definitions and the result we refer to Proposition 27.4 in \citet{Tomczak1989}:\\
\emph{If $B$ is a Banach space of type $p$ and cotype $q$, $1 < p \le 2 \le q < \infty$, and $\mathcal{X}$ is an $n$-dimensional subspace of $B$, then $d(\mathcal{X},\ell_2^n) \le C n^{1/p-1/q}$ for some constant $C$}.\\
Clearly, this bound is worse than $n^\alpha$, but it can be shown that this is the best possible asymptotic estimate for arbitrary $p,q$ provided $1/p-1/q < 1/2$. In the case $1/p-1/q \ge 1/2$ this proposition does not give anything new due to the fact that $\sqrt{n}$ is the maximal bound.
\end{remark}

\BP[Proof of Theorem~\ref{thm:BLAARLp}]
  Let $\mathcal{X} \subset L_p$ be the linear span of the input vectors $x_1,\ldots,x_T$.
  Let the dimension of $\mathcal{X}$ be $n$.
  By the Distance theorem
  there exists an isomorphism $U:\mathcal{X}\to \ell_2^n = \mathbb{R}^n$ such that
  \begin{equation*}
  \|U\|\|U^{-1}\| \le n^{|1/2-1/p|}=C.
  \end{equation*}
  Let the norms of the operator $U$ and of the inverse operator $U^{-1}$ be
  \begin{equation}\label{eq:IneqIsoNorms}
  \|U\| = \sup_{x \ne 0, x \in \mathcal{X}} \frac{\|U(x)\|}{\|x\|} = 1,
  \|U^{-1}\| = \sup_{r \ne 0, r \in \mathbb{R}^n} \frac{\|U^{-1}(r)\|}{\|r\|} \le C.
  \end{equation} Here we state the norm of the direct operator equals one since in the other case we can replace the operator $U$ by the operator $V = U / ||U||$ with unitary norm. The norm of the inverse operator then increases by $||U||$.

  By $r_i = U(x_i), i=1,\ldots,T$ we denote images of input vectors $x_i$ applying operator $U$. We apply the KAAR with the scalar product kernel to these images $r_i$ consequently. By formula~\eqref{eq:KAARbound} we get for any $g \in (\mathbb{R}^n)^*$ and $a > 0$ the loss of the learner at a step $T$ satisfies
  \begin{equation}\label{eq:BLAARinearg}
  L_T(\mathrm{BLAAR}) \le L_T(g) + a\|g\|_2^2 + Y^2T\frac{\max_{i=1,\ldots,T}{\|r_i\|_2^2}}{a},
  \end{equation}
  where the determinant of the positive definite matrix $I + \frac{1}{a}\widetilde{K}$ ($\widetilde{K}$ is the matrix of scalar products $(r_i,r_j)$) is bounded by the product of its diagonal elements~\citep[Chapter 2, Theorem 7]{Beckenbach1961} and the logarithm $\ln(1+x)$ is bounded by $x$ for $x \ge 0$.

  For any $f \in B^*$ we take $g: \mathbb{R}^n \to \mathbb{R}$ such that $g(r) := f(U^{-1}(r)), \forall r \in \mathbb{R}^n$. Since $U$ is an isomorphism, $U^{-1}(r) \in L_p$. The linearity of $g$ follows from the linearity of $U^{-1}$, so $g \in (\mathbb{R}^n)^*$. This means that $L_T(g)=L_T(f)$ because the values of $f$ and $g$ are equal on the corresponding vectors from $\mathcal{X}$ and $\mathbb{R}^n$.

  Let us consider any linear functional $h: \mathcal{X} \to \mathbb{R}$ on it such that $h(x) = f(x), \forall x \in \mathcal{X}$. Clearly,
  \begin{equation*}
  \|f\|_B = \sup_{\|x\|=1, x \in B} |f(x)| \ge \sup_{\|x\| = 1, x \in \mathcal{X}}|f(x)|
  = \sup_{\|x\|=1, x \in \mathcal{X}}|h(x)| = \|h\|_\mathcal{X}.
  \end{equation*}
  Then, the norm of $h$ can be lower estimated using \eqref{eq:IneqIsoNorms}:
  \begin{equation*}
  \|h\|_\mathcal{X} = \sup_{x \ne 0, x \in \mathcal{X}} \frac{|h(x)|}{\|x\|} = \sup_{r \ne 0, r \in \mathbb{R}^n} \frac{|g(r)|}{\|U^{-1}r\|}
  \ge \frac{1}{C}\sup_{r \ne 0, r \in \mathbb{R}^n} \frac{|g(r)|}{\|r\|} = \frac{1}{C}\|g\|.
  \end{equation*}
  On the other hand, we have $ \|r\| \le \|x\|$ for all $x \in \mathcal{X}, r=U(x)$. Thus $\|r_i\|^2 \le \|x_i\|^2, i=1,\ldots,T$ and $\|g\|^2 \le C^2\|h\|^2 \le C^2\|f\|^2$, so the theoretical bound \eqref{eq:BLAARinearg} transforms to
  \begin{equation*}
  L_T(\mathrm{BLAAR}) \le L_T(f) + aC^2\|f\|^2 + \frac{Y^2TX^2}{a}.
  \end{equation*}
  We can choose $a=\sqrt{T}/C$ and recall that $n$ in $C$ is the number of linearly independent input vectors among $x_1,\ldots,x_T$. Thus $n \le T$, and we get the bound~\eqref{eq:BoundBLAARLp}.
\EP

\subsection*{Derivation of Algorithm \ref{alg:lattices}}
The derivation of our algorithm is based on the proof of the Distance theorem for $B = L_p$ given in \cite{Lewis1978}.\\
\textbf{Step 2: The optimization task.} Let $\text{span}\{x_{r_1},\ldots,x_{r_n}\} = Z$, $\dim Z = n$ (we will further omit index $r$ of $x$-es). We take some basis $\phi_1,\ldots,\phi_n \in Z^*$. For an isomorphism $u: \ell_2^n = \mathbb{R}^n \to Z \subset L_{p}$ we define its determinant by $\det u = \det\{\phi_i(\gamma_j)\}_{ij}$, where $\gamma_i = u(e_i)$ for a unit vector basis $e_i \in \mathbb{R}^n$. Then we find such $u$ that $|\det u| \to \max$ and $\|\gamma_Z\|_p \le 1$, where $\gamma_Z = \sqrt{\sum_{i=1}^n|\gamma_i|^2}$ (the resulting isomorphism is the one where the infimum is attained in the Distance theorem). The maximum determinant is unique up to a constant depending on the choice of $\phi_i$. It is convenient to choose $\phi_i: \phi_i(x) = a_i$ for any $x = \sum_{i=1}^n a_i x_i \in Z$ (so $\{\phi_i\}_1^n$ is a biorthogonal system to $\{x_i\}_1^n$). Then $|\det u| = |\det \{c_{ij}\}_{ij}|$ for $\gamma_i = \sum_{j=1}^n c_{ij} x_j$.\\
\textbf{Step 3: The scalar product.} The scalar product \eqref{eq:scaltheta} is calculated by
\begin{equation*}
k_{ij} = \int_\Omega x_i x_j |\gamma_Z|^{p-2} dx, i,j=1,\ldots,T.
\end{equation*} and is equivalent to \eqref{eq:scaltheta} because $\delta_{ij} = n \int_\Omega \gamma_i \gamma_j |\gamma_Z|^{p'-2} dx, i,j=1,\ldots,T$ from the proof of Theorem 1 in \citet{Lewis1978}.

\subsection*{Proof of Corollary~\ref{cor:Sobolev}}
Sobolev spaces are proper Banach spaces \citep[see][Proposition 7(ii)]{Triebel2005}.
The identity mapping from them to $C(\mathbf{X})$ is bounded, so $c_{W_p^s} < \infty$. From the other hand, we impose a lattice structure on the Sobolev space using isomorphism \eqref{eq:convol}.
\BP
We represent $x_1,\ldots,x_T$ as elements $\alpha_i$ of a dual space $(W_p^{s})^*$. We take $\alpha_{i}(f) = \alpha_{x_i}(f) := f(x_i), \forall f \in W_p^s, i=1,\ldots,T$. In this setting we compete with the elements of $(W_p^s)^{**}$: for each $f \in W_p^s$ we take $g_f \in (W_p^s)^{**}$ such that by definition $g_f(\alpha) := \alpha(f)$ for any $\alpha \in (W_p^s)^{*}$. This changing of variables does not change the prediction error, since $f(x_i) = \alpha_i(f) = g_f(\alpha_i)$.

The isomorphism theorem states that there exists a linear isomorphism $U: L_p \to W_p^s$ between $L_p$ and $W_p^s$, such that $\|U\|\|U^{-1}\| < K$ for some constant $K$. We also denote $C_U = ||U|| =\sup_{\eta \in L_p} \frac{||U\eta||}{||\eta||}, C_{U^{-1}} = ||U^{-1}|| = \sup_{f \in W_p^s} \frac{||U^{-1}f||}{||f||}$. This isomorphism defines a dual isomorphism $U^*: (W_p^s)^* \to (L_p)^*$ by
\[
(U^* \alpha) (\eta) = \alpha (U \eta),
\]
where $\eta \in L_p$ and $\alpha \in (W_p^s)^*$. Clearly, $(U^* \alpha) (U^{-1}f) = \alpha(f), \forall f \in W_p^s$. We denote $\beta = U^* \alpha \in (L_p)^*$. Similarly, we have a correspondence $U^{**}: (W_p^s)^{**} \to (L_p)^{**}$ defined by
\[
(U^{**}g)(\beta) = g((U^*)^{-1}\beta),
\]
which gives us $h = U^{**}g \in (L_p)^{**}$ functions to compete with. After these replacements we have the same prediction error since for any $g \in (W_p^s)^{**}$ we get $h(\beta_i) = (U^{**}g)(\beta_i) = g((U^*)^{-1}\beta_i) = g(\alpha_i), i=1,\ldots,T$. The norm of a function increases by some constant:
\begin{equation*}
||h|| = \sup_\beta \frac{|h(\beta)|}{||\beta||} = \sup_\alpha \frac{|g(\alpha)|}{||U^* \alpha||}
\le \frac{|g(\alpha)|}{|\alpha(U\eta)|}||\eta|| = ||\eta|| \le C_{U^{-1}}\|f\|,
\end{equation*}
where $\eta = U^{-1}f$. The first inequality follows from the fact that $\|U^*\alpha\| = \sup_{\eta} \frac{|(U^*\alpha)(\eta)|}{\|\eta\|} \ge \frac{|(U^*\alpha)(\eta)|}{\|\eta\|} = \frac{|\alpha(U\eta)|}{\|\eta\|}, \forall \eta \in L_p$. To apply Theorem~\ref{thm:BLAARLp} we have to ensure $||\beta_i||$ is bounded. It holds since
\begin{equation*}
||\beta_i|| = \sup_{\eta} \frac{|(U^*\alpha_i)(\eta)|}{||\eta||} = \sup_f \frac{|f(x)|}{||U^{-1}f||}
\le \sup_f \frac{|f(x_i)|}{||f||} C_U \le c_{W_p^s} C_U, i=1,\ldots,T.
\end{equation*}

Applying Theorem~\ref{thm:BLAARLp} concludes the proof.
\EP

\end{document}